\documentclass[11pt,a4paper]{article}
\usepackage[hyperref]{acl2020}
\usepackage{times}
\usepackage{latexsym}
\usepackage{booktabs}

\usepackage{algorithm}
\usepackage[noend]{algpseudocode}
\usepackage{sidecap}
\algnewcommand\algorithmicinput{\textbf{Input:}}
\algnewcommand\algorithmicoutput{\textbf{Output:}}
\algnewcommand\Input{\item[\algorithmicinput]}%
\algnewcommand\Output{\item[\algorithmicoutput]}%
\usepackage{microtype}
\usepackage{graphicx}
\usepackage{amsmath}
\usepackage{subcaption}
\usepackage[export]{adjustbox}
\aclfinalcopy % Uncomment this line for the final submission
\DeclareMathOperator*{\argmin}{arg\,min}

\title{Towards Open Domain Event Trigger Identification using Adversarial Domain Adaptation}

\author{Aakanksha Naik \\
Carnegie Mellon University \\
    \texttt{anaik@cs.cmu.edu}
  \\\And
  Carolyn Ros\'e \\
Carnegie Mellon University\\
  \texttt{cprose@cs.cmu.edu}
  }

\date{}

\begin{document}
\setlength{\abovedisplayskip}{3pt}
\setlength{\belowdisplayskip}{3pt}

\maketitle
\begin{abstract}
We tackle the task of building supervised event trigger identification models which can generalize better across domains. Our work leverages the adversarial domain adaptation (ADA) framework to introduce domain-invariance. ADA uses adversarial training to construct representations that \emph{are predictive} for trigger identification, but \emph{not predictive} of the example's domain. It requires no labeled data from the target domain, making it completely unsupervised. Experiments with two domains (English literature and news) show that ADA leads to an average F1 score improvement of 3.9 on out-of-domain data. Our best performing model (BERT-A) reaches 44-49 F1 across both domains, using \textbf{no} labeled target data. Preliminary experiments reveal that finetuning on 1\% labeled data, followed by self-training leads to substantial improvement, reaching 51.5 and 67.2 F1 on literature and news respectively.\footnote{Our system is available at \url{https://github.com/aakanksha19/ODETTE}}
\end{abstract}

\section{Introduction}
Events are a key semantic phenomenon in natural language understanding. They embody a basic function of language: the ability to report happenings. Events are a basic building block for narratives across multiple domains such as news articles, stories and scientific abstracts, and are important for many downstream tasks such as question answering \cite{sauri2005evita} and summarization \cite{daniel2003sub}. Despite their utility, event extraction remains an onerous task. A major reason for this is that the notion of what counts as an ``event'' depends heavily on the domain and task at hand. For example, should a system which extracts events from doctor notes only focus on medical events (eg: symptoms, treatments), or also annotate lifestyle events (eg: dietary changes, exercise habits) which may have bearing on the patient's illness? To circumvent this, prior work has mainly focused on annotating specific categories of events \cite{grishman-sundheim-1996-message,doddington2004automatic,kim2008corpus} or narratives from specific domains \cite{pustejovsky2003timebank,sims-etal-2019-literary}. This has an important implication for supervised event extractors: they do not generalize to data from a different domain or containing different event types \cite{keith-etal-2017-identifying}. Conversely, event extractors that incorporate syntactic rule-based modules \cite{sauri2005evita,chambers-etal-2014-dense} tend to overgenerate, labeling most verbs and nouns as events. Achieving a balance between these extremes will help in building generalizable event extractors, a crucial problem since annotated training data may be expensive to obtain for every new domain.

Prior work has explored unsupervised \cite{huang-etal-2016-liberal,yuan2018open}, distantly supervised \cite{keith-etal-2017-identifying,chen-etal-2017-automatically,araki-mitamura-2018-open,zeng2018scale} and semi-supervised approaches \cite{liao-grishman-2010-filtered,huang-riloff-2012-bootstrapped,ferguson-etal-2018-semi}, which largely focus on automatically generating in-domain training data. In our work, we try to leverage annotated training data from other domains. Motivated by the hypothesis that events, despite being domain/ task-specific, often occur in similar contextual patterns, we try to inject lexical domain-invariance into supervised models, improving generalization, while not overpredicting events.  

Concretely, we focus on event trigger identification, which aims to identify triggers (words) that instantiate an event. For example, in ``John was born in Sussex'', \emph{born} is a trigger, invoking a \textsc{BIRTH} event. To introduce domain-invariance, we adopt the adversarial domain adaptation (ADA) framework \cite{ganin2015unsupervised} which constructs representations that \emph{are predictive} for trigger identification, but \emph{not predictive} of the example's domain, using adversarial training. This framework requires no labeled target domain data, making it completely unsupervised. Our experiments with two domains (English literature and news) show that ADA makes supervised models more robust on out-of-domain data, with an average F1 score improvement of 3.9, at no loss of in-domain performance. Our best performing model (BERT-A) reaches 44-49 F1 across both domains using \textbf{no} labeled data from the target domain. Further, preliminary experiments demonstrate that finetuning on 1\% labeled data, followed by self-training leads to substantial improvement, reaching 51.5 and 67.2 F1 on literature and news respectively. %\an{Should there be any connections to future ideas to explore here?}

%\section{ODETTE: \textbf{O}pen \textbf{D}omain \textbf{E}vent Trigger \textbf{T}agg\textbf{E}r}
\section{Approaching Open Domain Event Trigger Identification}
Throughout this work, we treat the task of event trigger identification as a token-level classification task. For each token in a sequence, we predict whether it is an event trigger. To ensure that our trigger identification model can transfer across domains, we leverage the adversarial domain adaptation (ADA) framework \cite{ganin2015unsupervised}, which has been used in several NLP tasks \cite{ganin2016domain,li2017end,liu-etal-2017-adversarial,chen-etal-2018-adversarial,shah-etal-2018-adversarial,yu2018modelling}. 

\subsection{Adversarial Domain Adaptation}
\begin{figure}[h]
    \centering
    \includegraphics[scale=0.25]{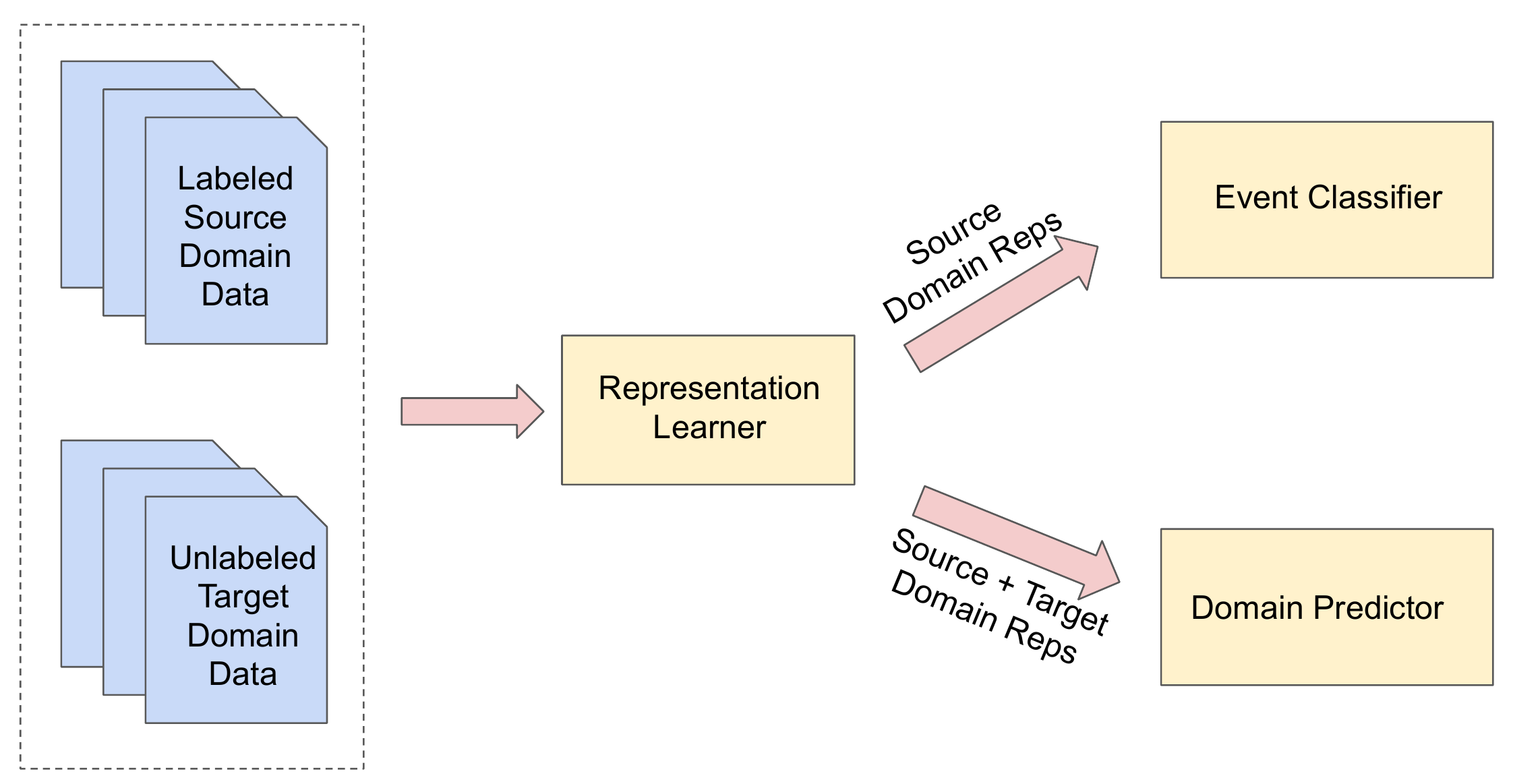}
    \caption{Adversarial Domain Adaptation Framework for Event Trigger Identification}
    \label{fig:ada}
\end{figure}

Figure \ref{fig:ada} gives an overview of the ADA framework for event trigger identification. It consists of three components: i) representation learner ($R$) ii) event classifier ($E$) and iii) domain predictor ($D$). The representation learner generates token-level representations, while the event classifier and domain predictor use these representations to identify event triggers and predict the domain to which the sequence belongs. The key idea is to train the representation learner to generate representations which \emph{are predictive} for trigger identification but \emph{not predictive} for domain prediction, making it more domain-invariant. A notable benefit here is that the only data we need from the target domain is unlabeled data. 

To ensure domain-invariant representation learning, ADA uses adversarial training. Assume that we have a labeled source domain dataset $D^s$ with examples $\{(x^s_1, e^s_1), ..., (x^s_n, e^s_n)\}$, where $x^s_i$ is the token sequence and $e^s_i$ is the sequence of event tags. We construct auxiliary dataset $D^a$ with examples $\{(x^a_1, d^a_1), ..., (x^a_n, d^a_n)\}$, where $x^a_i$ is the token sequence and $d^a_i$ is the domain label, using token sequences from $D^s$ and unlabeled target domain sentences. The representation learner $R$ maps a token sequence $x_i = (x_{i1},...,x_{ik})$ into token representations $h_i = (h_{i1},...,h_{ik})$. The event classifier $E$ maps representations $h_i = (h_{i1},...,h_{ik})$ to event tags $e_i = (e_{i1},...,e_{ik})$. The domain predictor $D$ creates a pooled representation $p_i = Pool(h_{i1},...,h_{ik})$ and maps it to domain label $d^a_i$. Given this setup, we apply an alternating optimization procedure. In the first step, we train the domain predictor using $D^a$, to optimize the following loss:\\
\vspace{-1.5em}
\begin{align*}
    \argmin_{D} \mathcal{L}(D(h^a_i), d^a_i)
\end{align*}
In the second step, we train the representation learner and event classifier using $D^s$ to optimize the following loss:\\
\vspace{-1em}
\begin{align*}
    \argmin_{R,E} \Bigg[ \sum_k \Big(\mathcal{L}(E(h^s_{ik}), e^s_{ik})\Big) - \lambda \mathcal{L}(D(h^s_i), d^s_i) \Bigg]
\end{align*}
$\mathcal{L}$ refers to the cross-entropy loss and $\lambda$ is a hyperparameter. In practice, the optimization in the above equation is performed using a gradient reversal layer (GRL) \cite{ganin2015unsupervised}. A GRL works as follows. During the forward pass, it acts as the identity, but during the backward pass it scales the gradients flowing through by $-\lambda$. We apply a GRL $g_\lambda$ before mapping the pooled representation to a domain label using $D$. This changes the optimization to:\\
\vspace{-1em}
\begin{align*}
    \argmin_{R,E} \Bigg[\mathcal{L}(D(g_\lambda(p^s_i)), d^s_i) + \sum_k \mathcal{L}(E(h^s_{ik}), e^s_{ik})  \Bigg]
\end{align*}
In our setup, the event classifier and domain predictors are MLP classifiers. For the representation learner, we experiment with several architectures.

\begin{table}[]
    \centering
    \small
    \begin{tabular}{rcc}
        \toprule \textbf{Statistic} & \textbf{LitBank} & \textbf{TimeBank} \\ \midrule
        \textbf{\#Docs} & 100 & 183 \\
        \textbf{\#Tokens} & 210,532 & 80,281 \\
        \textbf{\#Events} & 7849 & 8103 \\
        \textbf{Event Density} & 3.73\% & 10.10\% \\ \bottomrule
    \end{tabular}
    \caption{Dataset Statistics}
    \label{tab:datastats}
\end{table}

\subsection{Representation Learner Models}
We experiment with the following models:\footnote{Complete implementation details in the appendix}\\
\textbf{LSTM}: A unidirectional LSTM over tokens represented using word embeddings. \\
\textbf{BiLSTM}: A bidirectional LSTM over word embeddings to incorporate both left and right context. \\
\textbf{POS}: A BiLSTM over token representations constructed by concatenating word embeddings with embeddings corresponding to part-of-speech tags. This model explicitly introduces syntax. \\
% \textbf{DEP}:\\
% \textbf{CNN}: Inspired by the \cite{nguyen-grishman-2015-event}, we augment token-level representations from our BiLSTM with the outputs of a CNN over the entire sentence and positional embeddings. \\
\textbf{BERT}: A BiLSTM over contextual token representations extracted using BERT \cite{devlin-etal-2019-bert}, similar to the best-performing model on LitBank, reported by \newcite{sims-etal-2019-literary}.

\begin{table}[]
    \centering
    \small
    \begin{tabular}{rcccccc}
        \toprule \textbf{Model} & \multicolumn{3}{c}{\textbf{In-Domain}} & \multicolumn{3}{c}{\textbf{Out-of-Domain}}  \\ \cmidrule{2-7}
         & \textbf{P} & \textbf{R} & \textbf{F1} & \textbf{P} & \textbf{R} & \textbf{F1} \\ \midrule
        \textbf{LSTM} & 61.9 & 61.5 & 61.7 & 86.1 & 17.1 & 28.5 \\
        \textbf{LSTM-A} & 61.1 & 61.6 & 61.3 & 89.0 & 18.9 & 31.2 \\ \midrule
        \textbf{BiLSTM} & 64.5 & 61.7 & 63.1 & 91.8 & 14.4 & 24.9 \\
        \textbf{BiLSTM-A} & 66.1 & 62.8 & 64.4 & 92.9 & 18.5 & 30.9 \\ \midrule
        \textbf{POS} & 74.1 & 51.9 & 61.1 & 93.5 & 9.6 & 17.4 \\
        \textbf{POS-A} & 69.6 & 57.7 & 63.1 & 92.5 & 15.2 & 26.1 \\ \midrule
        % \textbf{DEP} & & & & & & \\
        % \textbf{DEP-A} & & & & & & \\ \midrule
        % \textbf{CNN} & & & & & & \\
        % \textbf{CNN-A} & & & & & & \\ \midrule
        \textbf{BERT} & 73.5 & 72.7 & 73.1 & \textbf{88.1} & 28.2 & 42.7 \\
        \textbf{BERT-A} & 71.9 & 71.3 & 71.6 & 85.0 & \textbf{35.0} & \textbf{49.6} \\ \bottomrule
    \end{tabular}
    \caption{Model performance on domain transfer experiments from LitBank to TimeBank. Presence of the -A suffix indicates that the model uses adversarial training.}
    \label{tab:slit}
\end{table}

\section{Experiments}
\subsection{Datasets}
In our experiments, we use the following datasets:\footnote{Unlike prior work, we cannot use the ACE-2005 dataset since it tags specific categories of events, whereas we focus on tagging \emph{all} possible events.}
\begin{itemize}
\vspace{-0.5em}
\setlength\itemsep{-0.5em}
    \item \textbf{LitBank} \cite{sims-etal-2019-literary}: 100 English literary texts with entity and event annotations.
    \item \textbf{TimeBank} \cite{pustejovsky2003timebank}: 183 English news articles containing annotations for events and temporal relations between them.
\end{itemize}
\vspace{-0.5em}
 Both datasets follow similar guidelines for event annotation, with an important distinction: LitBank does not annotate events which have not occurred (eg: future, hypothetical or negated events). To overcome this gap, we remove all such events from TimeBank using available metadata about event modality and tense. Table~\ref{tab:datastats} provides a brief overview of statistics for both datasets.

%\an{Should I move this to an appendix?}

\begin{table}[]
    \centering
    \small
    \begin{tabular}{rcccccc}
        \toprule \textbf{Model} & \multicolumn{3}{c}{\textbf{In-Domain}} & \multicolumn{3}{c}{\textbf{Out-of-Domain}}  \\ \cmidrule{2-7}
         & \textbf{P} & \textbf{R} & \textbf{F1} & \textbf{P} & \textbf{R} & \textbf{F1} \\ \midrule
        \textbf{LSTM} & 70.7 & 78.4 & 74.4 & 23.5 & 75.2 & 35.8 \\
        \textbf{LSTM-A} & 69.3 & 87.5 & 77.3 & 25.6 & 72.9 & 37.9 \\ \midrule
        \textbf{BiLSTM} & 75.4 & 76.3 & 75.9 & 27.6 & 68.8 & 39.4 \\
        \textbf{BiLSTM-A} & 74.2 & 79.4 & 76.7 & 26.3 & 72.0 & 38.6 \\ \midrule
        \textbf{POS} & 77.4 & 81.1 & 79.2 & 26.4 & 79.8 & 39.6 \\
        \textbf{POS-A} & 76.4 & 83.0 & 79.6 & 27.3 & 81.9 & 40.9 \\ \midrule
        % \textbf{DEP} & & & & & & \\
        % \textbf{DEP-A} & & & & & & \\ \midrule
        % \textbf{CNN} & & & & & & \\
        % \textbf{CNN-A} & & & & & & \\ \midrule
        \textbf{BERT} & 79.6 & 84.3 & 81.9 & 28.1 & \textbf{84.8} & 42.2 \\
        \textbf{BERT-A} & 79.8 & 85.6 & 82.6 & \textbf{30.3} & 80.8 &\textbf{ 44.1} \\ \bottomrule
    \end{tabular}
    \caption{Model performance on domain transfer experiments from TimeBank to LitBank. Presence of the -A suffix indicates that the model uses adversarial training.}
    \label{tab:snews}
\end{table}

\subsection{Results and Analysis}
Tables \ref{tab:slit} and \ref{tab:snews} present the results of our experiments. Table~\ref{tab:slit} shows the results when transferring from LitBank to TimeBank while Table~\ref{tab:snews} presents transfer results in the other direction. From Table~\ref{tab:slit} (transfer from LitBank to TimeBank), we see that ADA improves out-of-domain performance for all models, by 6.08 F1 on average. BERT-A performs best, reaching an F1 score of 49.6, using no labeled news data. Transfer experiments from TimeBank to LitBank (Table~\ref{tab:snews}) showcase similar trends, with only BiLSTM not showing improvement with ADA. For other models, ADA results in an average out-of-domain F1 score improvement of 1.77. BERT-A performs best, reaching an F1 score of 44.1. We also note that models transferred from LitBank to TimeBank have high precision, while models transferred in the other direction have high recall. We believe this difference stems from the disparity in event density across corpora (Table~\ref{tab:datastats}). Since event density in LitBank is much lower, models transferred from LitBank tend to be slightly conservative (high precision), while models transferred from TimeBank are less so (high recall).
%These results demonstrate that using ADA improves the generalizability of the representation learner, thus improving out-of-domain performance. 

% \begin{figure*}
% \centering
% \begin{subfigure}{.45\textwidth}
%   \centering
%   \includegraphics[scale=0.35]{acl2020-templates/figures/tb_finetune.png}
%   \caption{Improvement in model performance when finetuning on limited labeled training data from the target domain (TimeBank)}
%   \label{fig:tbfinetune}
% \end{subfigure}\hfill
% \begin{subfigure}{.45\textwidth}
%   \centering
%   \includegraphics[scale=0.35]{acl2020-templates/figures/lb_finetune.png}
%   \caption{Improvement in model performance when finetuning on limited labeled training data from the target domain (LitBank)}
%   \label{fig:lbfinetune}
% \end{subfigure}
% \label{fig:test}
% \end{figure*}
\begin{table}[]
    \centering
    \begin{tabular}{p{1.25cm}cp{4.5cm}}
      \toprule \textbf{Category} & \textbf{\%} & \textbf{Example} \\ \midrule
      \multicolumn{3}{c}{\textbf{TimeBank Improvements}}  \\ \midrule
        Finance & 54 & the \textbf{accord} was unanimously approved\\ 
        Political & 12 & the ukrainian parliament has already \textbf{ratified }it\\
        Reporting & 10 & from member station kqed , auncil martinez\textbf{ reports} \\
        Law & 10 & mr. antar was charged last month in a civil \textbf{suit}\\ \midrule
         \multicolumn{3}{c}{\textbf{LitBank Improvements}} \\  \midrule
        Archaic & 6 & his countenance became intolerably \textbf{fervid}\\
        Animal Actions & 6 & the dogs left off \textbf{barking} , and ran about every way \\
        Human Actions & 18 & a \textbf{nod} was the answer \\
        Literary & 14 & there \textbf{strikes} the ebony clock \\
         \bottomrule
    \end{tabular}
    \caption{Categorization of TimeBank and LitBank examples on which ADA shows improvement. Words in bold indicate events missed by BERT, but captured by BERT-A.}
    \label{tab:errors}
\end{table}

When transferring from LitBank to TimeBank, LSTM generalizes better than BiLSTM, which may be because BiLSTM has twice as many parameters making it more prone to overfitting. ADA gives a higher F1 boost with BiLSTM, indicating that it may be acting as a regularizer. Another interesting result is the poor performance of POS when transferring from LitBank to TimeBank. This might stem from the Stanford CoreNLP tagger (trained on news data) producing inaccurate tags for LitBank. Hence using automatically generated POS tags while training on LitBank does not produce reliable POS embeddings.

On average, ADA makes supervised models more robust on out-of-domain data, with an average F1 score improvement of 3.9, at no loss of in-domain performance.

\noindent
\textbf{What cases does ADA improve on?} To gain more insight into the improvements observed on using ADA, we perform a manual analysis of out-of-domain examples that BERT labels incorrectly, but BERT-A gets right. We carry out this analysis on 50 examples from TimeBank and LitBank each. We observe that an overwhelming number of cases from TimeBank use vocabulary in contexts unique to news (43/50 or 86\%). This includes examples of financial events, political events and reporting events that are rarer in literature, indicating that ADA manages to reduce event extraction models' reliance on lexical features. We make similar observations for LitBank though the proportion of improvement cases with literature-specific vocabulary is more modest (22/50 or 44\%). These cases include examples with archaic vocabulary, words that have a different meaning in literary contexts and human/ animal actions, which are not common in news. Table~\ref{tab:errors} presents a detailed breakdown of these cases, along with examples.\footnote{This table does not include generic improvement cases (i.e. no domain-specific vocabulary used), which formed 14\% and 56\% of improvement cases in TimeBank and LitBank.}

\begin{figure}
\setlength{\belowcaptionskip}{-5pt}
    \centering
    \includegraphics[scale=0.3]{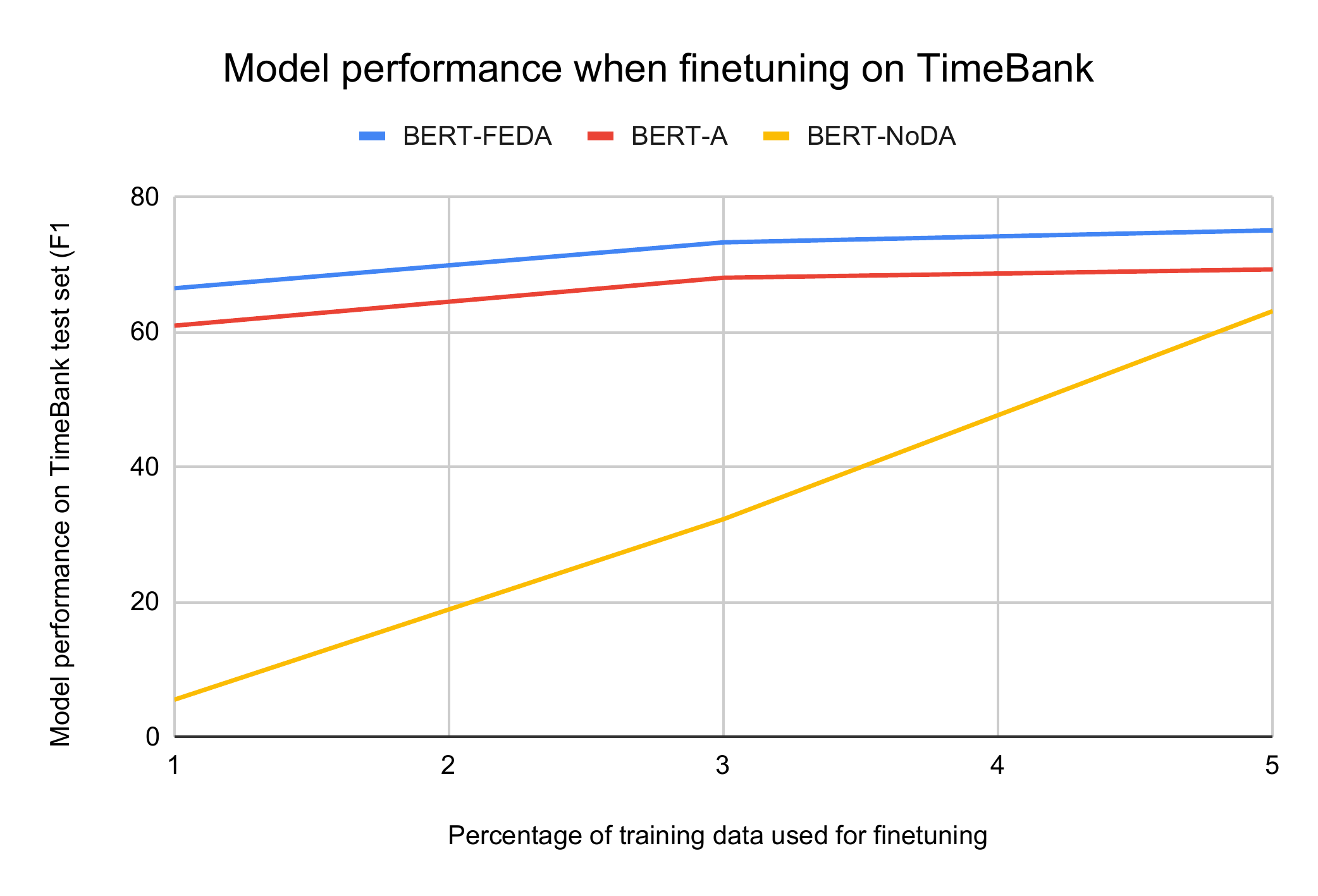}
    \caption{Improvement in model performance when finetuning on labeled training data from TimeBank}
    \label{fig:tbfinetune}
\end{figure}

\begin{figure}
    \centering
    \includegraphics[scale=0.3]{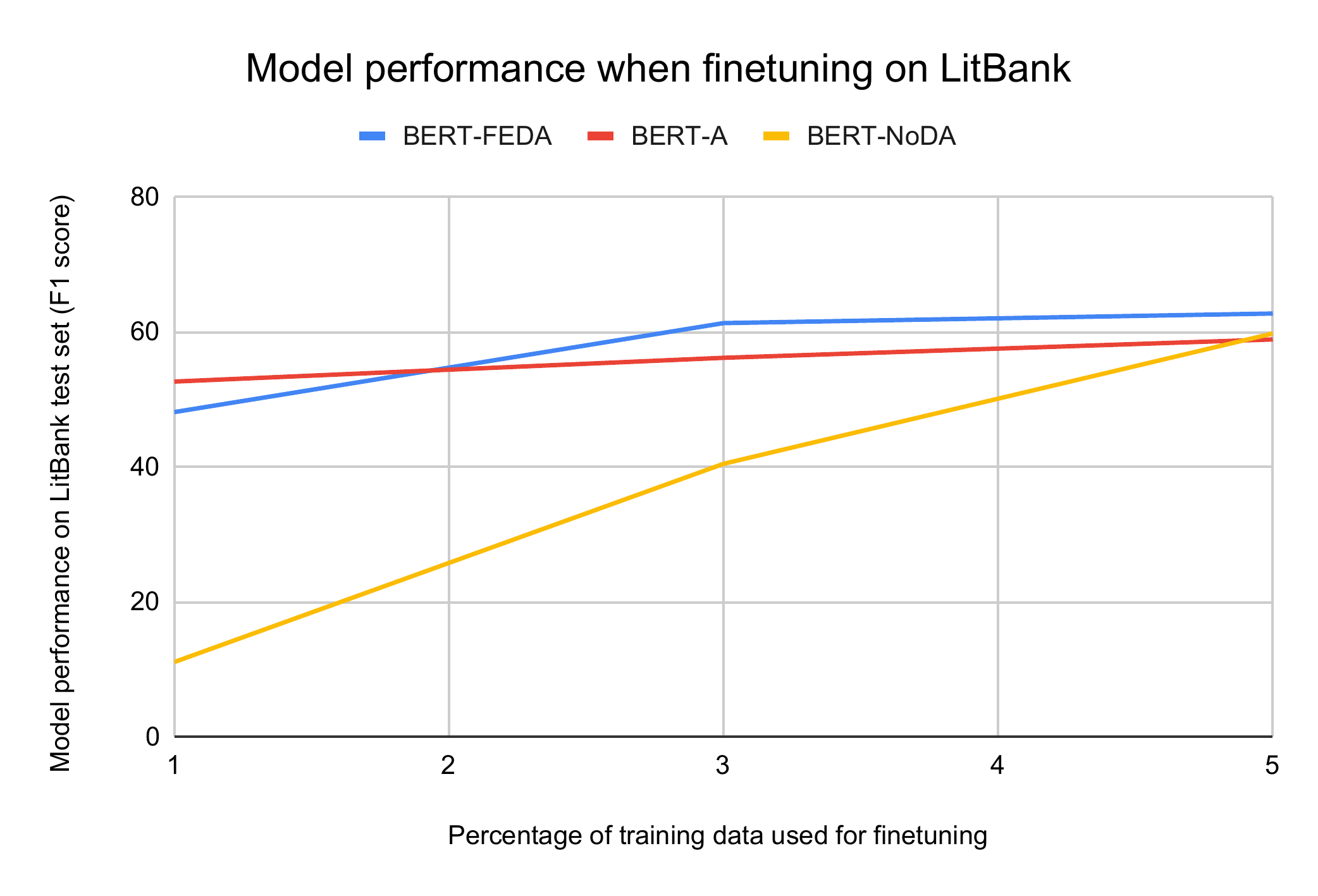}
    \caption{Improvement in model performance when finetuning on labeled training data from LitBank}
    \label{fig:lbfinetune}
\end{figure}

\section{Incorporating Minimal Labeled Data}
\textbf{Finetuning on labeled data:} We run finetuning experiments to study improvement in model performance on incorporating small amounts of labeled target domain data. For both domains, we finetune BERT-A, slowly increasing the percentage of labeled data used from 1\%-5\%.\footnote{We run these experiments 5 times with different random subsets and average performance across all runs.} We compare BERT-A with two other models. The first model is naive BERT with no domain adaptation (BERT-NoDA). The second model is a BERT model trained via supervised domain adaptation (BERT-FEDA), which we use as an indicator of ceiling performance. The supervised domain adaptation method we use is the neural modification of frustratingly easy domain adaptation developed in \newcite{kim-etal-2016-frustratingly}. Frustratingly easy domain adaptation \cite{daume-iii-2007-frustratingly} uses a feature augmentation strategy to improve performance when annotated data from both source and target domains is available. This algorithm simply duplicates input features 3 times, creating a source-specific, target-specific and general version of each feature. For source data, only the source-specific and general features are active, while only the target-specific and general features are active for target data. The neural modification works by duplicating the feature extractor module, which is the BiLSTM in our case. 

Figures \ref{fig:tbfinetune} and \ref{fig:lbfinetune} present the results of these experiments. Performance of all models steadily improves with more data, but BERT-A starts with a much higher F1 score than BERT-NoDA, demonstrating that ADA boosts performance when little annotated training data is available. Performance increase of BERT-NoDA is suprisingly rapid, especially on LitBank. However, it is worth noting that 5\% of the LitBank training set is $\sim$10,000 tokens, which is a substantial amount to annotate. Therefore, BERT-A beats BERT-NoDA on sample efficiency. We can also see that BERT-A does not do much worse than BERT-FEDA, which performs supervised adaptation.

\noindent
\textbf{Using BERT-A to provide weak supervision:} We run further experiments to determine whether finetuned BERT-A can be leveraged for self-training \cite{yarowsky-1995-unsupervised,riloff2003learning}. Self-training creates a teacher model from labeled data, which is then used to label a large amount of unlabeled data. Both labeled and unlabeled datasets are jointly used to train a student model. Algorithm \ref{alg:selftrain} gives a quick overview of our self-training procedure. We use 1\% of the training data as $\mathcal{D}^l$, with the remaining 99\% used as  $\mathcal{D}^u$. BERT-A acts as $\mathcal{T}$, while $\mathcal{S}$ is a vanilla BERT model. Table~\ref{tab:selfsup} shows the results of self-training on both domains. Self-training improves model performance by nearly 7 F1 points on average. Increase on TimeBank is much higher which may be due to the high precision-low recall tendency of the teacher model.

\begin{algorithm}[h]
\caption{SelfTrain($\mathcal{D}^l,\mathcal{D}^u,\mathcal{T}$)}
\label{alg:selftrain}
\begin{algorithmic}[1]
\Input Teacher Model $\mathcal{T}$, Labeled Data $\mathcal{D}^l = \{(x^l_1, e^l_1),... ,(x^l_m,x^l_m)\}$, Unlabeled Data $\mathcal{D}^u = \{x^u_1,...x^u_n\}$, 
\Output Trained Student Model $\mathcal{S}$
\State Finetune the teacher model $\mathcal{T}$ by minimizing cross-entropy loss on labeled data 
\begin{align*}
    \frac{1}{m} \sum_{i=1}^m \mathcal{L}(\mathcal{T}(x^l_i), e^l_i)
\end{align*}
\State Generate labels $\{e^u_1,...,e^u_n\}$ for unlabeled data $\mathcal{D}^u$ using $\mathcal{T}$
\State Train a student model $\mathcal{S}$ by minimizing cross-entropy loss on both datasets $\mathcal{D}^l, \mathcal{D}^u$
\begin{align*}
    \frac{1}{m} \sum_{i=1}^m \mathcal{L}(\mathcal{S}(x^l_i), e^l_i) + \frac{1}{n} \sum_{i=1}^n \mathcal{L}(\mathcal{S}(x^u_i), e^u_i)
\end{align*}
\State Iterative training: Repeat step 2 using updated student model $\mathcal{S}$
\end{algorithmic}
\end{algorithm}

\begin{table}[]
    \centering
    \begin{tabular}{rccc}
        \toprule \textbf{Dataset} & \textbf{P} & \textbf{R} & \textbf{F1}  \\ \midrule
        \textbf{TimeBank} & 68.9 & 65.5 & 67.2 \\
        \textbf{LitBank} & 40.3 & 71.5 & 51.5 \\ \bottomrule
    \end{tabular}
    \caption{Model performance on both domains in the self-training paradigm}
    \label{tab:selfsup}
\end{table}

\section{Conclusion}
In this work, we tackled the task of building generalizable supervised event trigger identification models using adversarial domain adaptation (ADA) to introduce domain-invariance. Our experiments with two domains (English literature and news) showed that ADA made supervised models more robust on out-of-domain data, with an average F1 score improvement of 3.9. Our best performing model (BERT-A) was able to reach 44-49 F1 across both domains using \textbf{no} labeled target domain data. Preliminary experiments showed that finetuning BERT-A on 1\% labeled data, followed by self-training led to substantial improvement, reaching 51.5 and 67.2 F1 on literature and news respectively. While these results are encouraging, we are yet to match supervised in-domain model performance. Future directions to explore include incorporating noise-robust training procedures \cite{goldberger2016training} and example weighting \cite{dehghani2018fidelity} during self-training, and exploring lexical alignment methods from literature on learning cross-lingual embeddings.
%leveraging domain adapted models to generate noisy automatically labeled training data for semi-supervised learning, and using self-training to further improve the BERT-A model

\section*{Acknowledgements}
This work was supported by the University of Pittsburgh Medical Center (UPMC) and Abridge AI Inc through the Center for Machine Learning and Health at Carnegie Mellon University. The authors would like to thank the anonymous reviewers for their helpful feedback on this work.

\bibliography{acl2020}

\begin{thebibliography}{31}
\expandafter\ifx\csname natexlab\endcsname\relax\def\natexlab#1{#1}\fi

\bibitem[{Araki and Mitamura(2018)}]{araki-mitamura-2018-open}
Jun Araki and Teruko Mitamura. 2018.
\newblock \href {https://www.aclweb.org/anthology/C18-1075} {Open-domain event
  detection using distant supervision}.
\newblock In \emph{Proceedings of the 27th International Conference on
  Computational Linguistics}, pages 878--891, Santa Fe, New Mexico, USA.
  Association for Computational Linguistics.

\bibitem[{Chambers et~al.(2014)Chambers, Cassidy, McDowell, and
  Bethard}]{chambers-etal-2014-dense}
Nathanael Chambers, Taylor Cassidy, Bill McDowell, and Steven Bethard. 2014.
\newblock \href {https://doi.org/10.1162/tacl_a_00182} {Dense event ordering
  with a multi-pass architecture}.
\newblock \emph{Transactions of the Association for Computational Linguistics},
  2:273--284.

\bibitem[{Chen et~al.(2018)Chen, Sun, Athiwaratkun, Cardie, and
  Weinberger}]{chen-etal-2018-adversarial}
Xilun Chen, Yu~Sun, Ben Athiwaratkun, Claire Cardie, and Kilian Weinberger.
  2018.
\newblock \href {https://doi.org/10.1162/tacl_a_00039} {Adversarial deep
  averaging networks for cross-lingual sentiment classification}.
\newblock \emph{Transactions of the Association for Computational Linguistics},
  6:557--570.

\bibitem[{Chen et~al.(2017)Chen, Liu, Zhang, Liu, and
  Zhao}]{chen-etal-2017-automatically}
Yubo Chen, Shulin Liu, Xiang Zhang, Kang Liu, and Jun Zhao. 2017.
\newblock \href {https://doi.org/10.18653/v1/P17-1038} {Automatically labeled
  data generation for large scale event extraction}.
\newblock In \emph{Proceedings of the 55th Annual Meeting of the Association
  for Computational Linguistics (Volume 1: Long Papers)}, pages 409--419,
  Vancouver, Canada. Association for Computational Linguistics.

\bibitem[{Daniel et~al.(2003)Daniel, Radev, and Allison}]{daniel2003sub}
Naomi Daniel, Dragomir Radev, and Timothy Allison. 2003.
\newblock \href {https://www.aclweb.org/anthology/W03-0502} {Sub-event based
  multi-document summarization}.
\newblock In \emph{Proceedings of the {HLT}-{NAACL} 03 Text Summarization
  Workshop}, pages 9--16.

\bibitem[{Daum{\'e}~III(2007)}]{daume-iii-2007-frustratingly}
Hal Daum{\'e}~III. 2007.
\newblock \href {https://www.aclweb.org/anthology/P07-1033} {Frustratingly easy
  domain adaptation}.
\newblock In \emph{Proceedings of the 45th Annual Meeting of the Association of
  Computational Linguistics}, pages 256--263, Prague, Czech Republic.
  Association for Computational Linguistics.

\bibitem[{Dehghani et~al.(2018)Dehghani, Mehrjou, Gouws, Kamps, and
  Sch{\"{o}}lkopf}]{dehghani2018fidelity}
Mostafa Dehghani, Arash Mehrjou, Stephan Gouws, Jaap Kamps, and Bernhard
  Sch{\"{o}}lkopf. 2018.
\newblock \href {https://openreview.net/forum?id=B1X0mzZCW} {Fidelity-weighted
  learning}.
\newblock In \emph{6th International Conference on Learning Representations,
  {ICLR} 2018, Vancouver, BC, Canada, April 30 - May 3, 2018, Conference Track
  Proceedings}. OpenReview.net.

\bibitem[{Devlin et~al.(2019)Devlin, Chang, Lee, and
  Toutanova}]{devlin-etal-2019-bert}
Jacob Devlin, Ming-Wei Chang, Kenton Lee, and Kristina Toutanova. 2019.
\newblock \href {https://doi.org/10.18653/v1/N19-1423} {{BERT}: Pre-training of
  deep bidirectional transformers for language understanding}.
\newblock In \emph{Proceedings of the 2019 Conference of the North {A}merican
  Chapter of the Association for Computational Linguistics: Human Language
  Technologies, Volume 1 (Long and Short Papers)}, pages 4171--4186,
  Minneapolis, Minnesota. Association for Computational Linguistics.

\bibitem[{Doddington et~al.(2004)Doddington, Mitchell, Przybocki, Ramshaw,
  Strassel, and Weischedel}]{doddington2004automatic}
George Doddington, Alexis Mitchell, Mark Przybocki, Lance Ramshaw, Stephanie
  Strassel, and Ralph Weischedel. 2004.
\newblock \href {http://www.lrec-conf.org/proceedings/lrec2004/pdf/5.pdf} {The
  automatic content extraction ({ACE}) program {--} tasks, data, and
  evaluation}.
\newblock In \emph{Proceedings of the Fourth International Conference on
  Language Resources and Evaluation ({LREC}{'}04)}, Lisbon, Portugal. European
  Language Resources Association (ELRA).

\bibitem[{Ferguson et~al.(2018)Ferguson, Lockard, Weld, and
  Hajishirzi}]{ferguson-etal-2018-semi}
James Ferguson, Colin Lockard, Daniel Weld, and Hannaneh Hajishirzi. 2018.
\newblock \href {https://doi.org/10.18653/v1/N18-2058} {Semi-supervised event
  extraction with paraphrase clusters}.
\newblock In \emph{Proceedings of the 2018 Conference of the North {A}merican
  Chapter of the Association for Computational Linguistics: Human Language
  Technologies, Volume 2 (Short Papers)}, pages 359--364, New Orleans,
  Louisiana. Association for Computational Linguistics.

\bibitem[{Ganin and Lempitsky(2015)}]{ganin2015unsupervised}
Yaroslav Ganin and Victor~S. Lempitsky. 2015.
\newblock \href {http://proceedings.mlr.press/v37/ganin15.html} {Unsupervised
  domain adaptation by backpropagation}.
\newblock In \emph{Proceedings of the 32nd International Conference on Machine
  Learning, {ICML} 2015, Lille, France, 6-11 July 2015}, volume~37 of
  \emph{{JMLR} Workshop and Conference Proceedings}, pages 1180--1189.
  JMLR.org.

\bibitem[{Ganin et~al.(2016)Ganin, Ustinova, Ajakan, Germain, Larochelle,
  Laviolette, Marchand, and Lempitsky}]{ganin2016domain}
Yaroslav Ganin, Evgeniya Ustinova, Hana Ajakan, Pascal Germain, Hugo
  Larochelle, Fran{\c{c}}ois Laviolette, Mario Marchand, and Victor~S.
  Lempitsky. 2016.
\newblock \href {http://jmlr.org/papers/v17/15-239.html} {Domain-adversarial
  training of neural networks}.
\newblock \emph{J. Mach. Learn. Res.}, 17:59:1--59:35.

\bibitem[{Goldberger and Ben{-}Reuven(2017)}]{goldberger2016training}
Jacob Goldberger and Ehud Ben{-}Reuven. 2017.
\newblock \href {https://openreview.net/forum?id=H12GRgcxg} {Training deep
  neural-networks using a noise adaptation layer}.
\newblock In \emph{5th International Conference on Learning Representations,
  {ICLR} 2017, Toulon, France, April 24-26, 2017, Conference Track
  Proceedings}. OpenReview.net.

\bibitem[{Grishman and Sundheim(1996)}]{grishman-sundheim-1996-message}
Ralph Grishman and Beth Sundheim. 1996.
\newblock \href {https://www.aclweb.org/anthology/C96-1079} {Message
  understanding conference- 6: A brief history}.
\newblock In \emph{{COLING} 1996 Volume 1: The 16th International Conference on
  Computational Linguistics}.

\bibitem[{Huang et~al.(2016)Huang, Cassidy, Feng, Ji, Voss, Han, and
  Sil}]{huang-etal-2016-liberal}
Lifu Huang, Taylor Cassidy, Xiaocheng Feng, Heng Ji, Clare~R. Voss, Jiawei Han,
  and Avirup Sil. 2016.
\newblock \href {https://doi.org/10.18653/v1/P16-1025} {Liberal event
  extraction and event schema induction}.
\newblock In \emph{Proceedings of the 54th Annual Meeting of the Association
  for Computational Linguistics (Volume 1: Long Papers)}, pages 258--268,
  Berlin, Germany. Association for Computational Linguistics.

\bibitem[{Huang and Riloff(2012)}]{huang-riloff-2012-bootstrapped}
Ruihong Huang and Ellen Riloff. 2012.
\newblock \href {https://www.aclweb.org/anthology/E12-1029} {Bootstrapped
  training of event extraction classifiers}.
\newblock In \emph{Proceedings of the 13th Conference of the {E}uropean Chapter
  of the Association for Computational Linguistics}, pages 286--295, Avignon,
  France. Association for Computational Linguistics.

\bibitem[{Keith et~al.(2017)Keith, Handler, Pinkham, Magliozzi, McDuffie, and
  O{'}Connor}]{keith-etal-2017-identifying}
Katherine Keith, Abram Handler, Michael Pinkham, Cara Magliozzi, Joshua
  McDuffie, and Brendan O{'}Connor. 2017.
\newblock \href {https://doi.org/10.18653/v1/D17-1163} {Identifying civilians
  killed by police with distantly supervised entity-event extraction}.
\newblock In \emph{Proceedings of the 2017 Conference on Empirical Methods in
  Natural Language Processing}, pages 1547--1557, Copenhagen, Denmark.
  Association for Computational Linguistics.

\bibitem[{Kim et~al.(2008)Kim, Ohta, and Tsujii}]{kim2008corpus}
Jin{-}Dong Kim, Tomoko Ohta, and Jun'ichi Tsujii. 2008.
\newblock \href {https://doi.org/10.1186/1471-2105-9-10} {Corpus annotation for
  mining biomedical events from literature}.
\newblock \emph{{BMC} Bioinform.}, 9.

\bibitem[{Kim et~al.(2016)Kim, Stratos, and
  Sarikaya}]{kim-etal-2016-frustratingly}
Young-Bum Kim, Karl Stratos, and Ruhi Sarikaya. 2016.
\newblock \href {https://www.aclweb.org/anthology/C16-1038} {Frustratingly easy
  neural domain adaptation}.
\newblock In \emph{Proceedings of {COLING} 2016, the 26th International
  Conference on Computational Linguistics: Technical Papers}, pages 387--396,
  Osaka, Japan. The COLING 2016 Organizing Committee.

\bibitem[{Li et~al.(2017)Li, Zhang, Wei, Wu, and Yang}]{li2017end}
Zheng Li, Yu~Zhang, Ying Wei, Yuxiang Wu, and Qiang Yang. 2017.
\newblock \href {https://doi.org/10.24963/ijcai.2017/311} {End-to-end
  adversarial memory network for cross-domain sentiment classification}.
\newblock In \emph{Proceedings of the Twenty-Sixth International Joint
  Conference on Artificial Intelligence, {IJCAI} 2017, Melbourne, Australia,
  August 19-25, 2017}, pages 2237--2243. ijcai.org.

\bibitem[{Liao and Grishman(2010)}]{liao-grishman-2010-filtered}
Shasha Liao and Ralph Grishman. 2010.
\newblock \href {https://www.aclweb.org/anthology/C10-1077} {Filtered ranking
  for bootstrapping in event extraction}.
\newblock In \emph{Proceedings of the 23rd International Conference on
  Computational Linguistics (Coling 2010)}, pages 680--688, Beijing, China.
  Coling 2010 Organizing Committee.

\bibitem[{Liu et~al.(2017)Liu, Qiu, and Huang}]{liu-etal-2017-adversarial}
Pengfei Liu, Xipeng Qiu, and Xuanjing Huang. 2017.
\newblock \href {https://doi.org/10.18653/v1/P17-1001} {Adversarial multi-task
  learning for text classification}.
\newblock In \emph{Proceedings of the 55th Annual Meeting of the Association
  for Computational Linguistics (Volume 1: Long Papers)}, pages 1--10,
  Vancouver, Canada. Association for Computational Linguistics.

\bibitem[{Pustejovsky et~al.(2003)Pustejovsky, Hanks, Sauri, See, Gaizauskas,
  Setzer, Radev, Sundheim, Day, Ferro et~al.}]{pustejovsky2003timebank}
James Pustejovsky, Patrick Hanks, Roser Sauri, Andrew See, Robert Gaizauskas,
  Andrea Setzer, Dragomir Radev, Beth Sundheim, David Day, Lisa Ferro, et~al.
  2003.
\newblock The timebank corpus.
\newblock In \emph{Corpus linguistics}, volume 2003, page~40. Lancaster, UK.

\bibitem[{Riloff and Wiebe(2003)}]{riloff2003learning}
Ellen Riloff and Janyce Wiebe. 2003.
\newblock \href {https://www.aclweb.org/anthology/W03-1014} {Learning
  extraction patterns for subjective expressions}.
\newblock In \emph{Proceedings of the 2003 Conference on Empirical Methods in
  Natural Language Processing}, pages 105--112.

\bibitem[{Saur{\'\i} et~al.(2005)Saur{\'\i}, Knippen, Verhagen, and
  Pustejovsky}]{sauri2005evita}
Roser Saur{\'\i}, Robert Knippen, Marc Verhagen, and James Pustejovsky. 2005.
\newblock \href {https://www.aclweb.org/anthology/H05-1088} {{E}vita: A robust
  event recognizer for {QA} systems}.
\newblock In \emph{Proceedings of Human Language Technology Conference and
  Conference on Empirical Methods in Natural Language Processing}, pages
  700--707, Vancouver, British Columbia, Canada. Association for Computational
  Linguistics.

\bibitem[{Shah et~al.(2018)Shah, Lei, Moschitti, Romeo, and
  Nakov}]{shah-etal-2018-adversarial}
Darsh Shah, Tao Lei, Alessandro Moschitti, Salvatore Romeo, and Preslav Nakov.
  2018.
\newblock \href {https://doi.org/10.18653/v1/D18-1131} {Adversarial domain
  adaptation for duplicate question detection}.
\newblock In \emph{Proceedings of the 2018 Conference on Empirical Methods in
  Natural Language Processing}, pages 1056--1063, Brussels, Belgium.
  Association for Computational Linguistics.

\bibitem[{Sims et~al.(2019)Sims, Park, and Bamman}]{sims-etal-2019-literary}
Matthew Sims, Jong~Ho Park, and David Bamman. 2019.
\newblock \href {https://doi.org/10.18653/v1/P19-1353} {Literary event
  detection}.
\newblock In \emph{Proceedings of the 57th Annual Meeting of the Association
  for Computational Linguistics}, pages 3623--3634, Florence, Italy.
  Association for Computational Linguistics.

\bibitem[{Yarowsky(1995)}]{yarowsky-1995-unsupervised}
David Yarowsky. 1995.
\newblock \href {https://doi.org/10.3115/981658.981684} {Unsupervised word
  sense disambiguation rivaling supervised methods}.
\newblock In \emph{33rd Annual Meeting of the Association for Computational
  Linguistics}, pages 189--196, Cambridge, Massachusetts, USA. Association for
  Computational Linguistics.

\bibitem[{Yu et~al.(2018)Yu, Qiu, Jiang, Huang, Song, Chu, and
  Chen}]{yu2018modelling}
Jianfei Yu, Minghui Qiu, Jing Jiang, Jun Huang, Shuangyong Song, Wei Chu, and
  Haiqing Chen. 2018.
\newblock \href {https://doi.org/10.1145/3159652.3159685} {Modelling domain
  relationships for transfer learning on retrieval-based question answering
  systems in e-commerce}.
\newblock In \emph{Proceedings of the Eleventh {ACM} International Conference
  on Web Search and Data Mining, {WSDM} 2018, Marina Del Rey, CA, USA, February
  5-9, 2018}, pages 682--690. {ACM}.

\bibitem[{Yuan et~al.(2018)Yuan, Ren, He, Zhang, Geng, Huang, Ji, Lin, and
  Han}]{yuan2018open}
Quan Yuan, Xiang Ren, Wenqi He, Chao Zhang, Xinhe Geng, Lifu Huang, Heng Ji,
  Chin{-}Yew Lin, and Jiawei Han. 2018.
\newblock \href {https://doi.org/10.1145/3269206.3271674} {Open-schema event
  profiling for massive news corpora}.
\newblock In \emph{Proceedings of the 27th {ACM} International Conference on
  Information and Knowledge Management, {CIKM} 2018, Torino, Italy, October
  22-26, 2018}, pages 587--596. {ACM}.

\bibitem[{Zeng et~al.(2018)Zeng, Feng, Ma, Wang, Yan, Shi, and
  Zhao}]{zeng2018scale}
Ying Zeng, Yansong Feng, Rong Ma, Zheng Wang, Rui Yan, Chongde Shi, and Dongyan
  Zhao. 2018.
\newblock \href
  {https://www.aaai.org/ocs/index.php/AAAI/AAAI18/paper/view/16119} {Scale up
  event extraction learning via automatic training data generation}.
\newblock In \emph{Proceedings of the Thirty-Second {AAAI} Conference on
  Artificial Intelligence, (AAAI-18), the 30th innovative Applications of
  Artificial Intelligence (IAAI-18), and the 8th {AAAI} Symposium on
  Educational Advances in Artificial Intelligence (EAAI-18), New Orleans,
  Louisiana, USA, February 2-7, 2018}, pages 6045--6052. {AAAI} Press.

\end{thebibliography}
\bibliographystyle{acl_natbib}

\appendix
\section*{Appendix}
\section{Implementation Details}
\label{sec:supplemental}
All models are implemented in PyTorch. We use 300-dimensional GloVe embeddings while training on TimeBank and 100-dimensional Word2Vec embeddings trained on Project Gutenberg texts (similar to \cite{sims-etal-2019-literary}) while training on LitBank. Both source and target domains share a common vocabulary and embedding layer which is not finetuned during the training process. All LSTM models use a hidden size of 100, with an input dropout of 0.5. The POS model uses 50-dimensional embeddings for POS tags which are randomly initialized and finetuned during training. The BERT model uses the uncased variant of BERT-Base for feature extraction. We generate token representations by running BERT-Base and concatenating the outputs of the model's last 4 hidden layers. The event classifier is a single-layer 100-dimensional MLP. For the adversarial training setup, we experiment with values from [0.1,0.2,0.5,1.0,2.0,5.0] for the hyperparameter $\lambda$. The domain predictor (adversary) is a 3-layer MLP with each layer having a dimensionality of 100 and ReLU activations between layers. We train all models with a batch size of 16 and use the Adam optimizer with default learning rate settings. Models are trained for 1000 epochs, with early stopping. For finetuning experiments, we train for 10 epochs. 

\end{document}